\documentclass[]{bytedance_seed}

\usepackage[toc,page,header]{appendix}

\usepackage{hyperref}
\usepackage{url}
\usepackage{algorithm}
\usepackage{algorithmic}
\usepackage{enumitem}
\usepackage{fvextra}
\usepackage{multirow}
\usepackage{booktabs}
\usepackage[table]{xcolor}
\usepackage{wrapfig}
\usepackage{graphicx} 
\usepackage{fvextra}
\usepackage{url}
\usepackage{booktabs}
\usepackage{pifont}
\usepackage{tabularx}
\usepackage{amssymb}
\usepackage{longtable}
\usepackage{multirow}
\usepackage{multicol}
\usepackage{wrapfig}

\usepackage{minitoc}
\usepackage{tikz}
\usetikzlibrary{decorations.fractals}
\usetikzlibrary{lindenmayersystems}

\definecolor{kellygreen}{rgb}{0.3, 0.73, 0.09}
\definecolor{alizarin}{rgb}{0.82, 0.1, 0.26}
\newcommand{\cmark}{{\color{kellygreen} \ding{51}}}
\newcommand{\xmark}{{\color{alizarin} \ding{55}}}

\newcommand{\benchmark}{LongReason}

\newlength\savewidth\newcommand\shline{\noalign{\global\savewidth\arrayrulewidth
  \global\arrayrulewidth 1pt}\hline\noalign{\global\arrayrulewidth\savewidth}}

\pgfdeclarelindenmayersystem{Koch}{
  \rule{F -> F+F--F+F}
}
\pgfdeclarelindenmayersystem{Plant}{
  \rule{X -> F-[[X]+X]+F[+FX]-X}
  \rule{F -> FF}
}
\pgfdeclarelindenmayersystem{Dragon}{
  \rule{X -> X+YF+}
  \rule{Y -> -FX-Y}
}

\newcommand{\decoraten}[2]{%
  \ifnum#1>0
    decorate{ \decoraten{\numexpr#1-1\relax}{#2} }%
  \else
    #2%
  \fi}

\title{LongReason: A Synthetic Long-Context Reasoning Benchmark via Context Expansion}

\author[1,2,*,\dagger]{Zhan Ling}
\author[1]{Kang Liu}
\author[1,3,*]{Kai Yan}
\author[1,*]{Yifan Yang}
\author[1]{Weijian Lin}
\author[1]{Ting-Han Fan}
\author[1]{Lingfeng Shen}
\author[1]{Zhengyin Du}
\author[1, \dagger]{Jiecao Chen}

\affiliation[1]{ByteDance Seed}
\affiliation[2]{UC San Diego}
\affiliation[3]{University of Illinois Urbana-Champaign}

\contribution[*]{Work done at ByteDance Seed}
\contribution[\dagger]{Corresponding authors}

\abstract{
Large language models~(LLMs) have demonstrated remarkable progress in understanding long-context inputs. However, benchmarks for evaluating the long-context reasoning abilities of LLMs fall behind the pace. Existing benchmarks often focus on a narrow range of tasks or those that do not demand complex reasoning. To address this gap and enable a more comprehensive evaluation of the long-context reasoning capabilities of current LLMs, we propose a new synthetic benchmark, \textbf{\benchmark{}}, which is constructed by synthesizing long-context reasoning questions from a varied set of short-context reasoning questions through context expansion. \benchmark{} consists of 794 multiple-choice reasoning questions with diverse reasoning patterns across three task categories:~reading comprehension, logical inference, and mathematical word problems. We evaluate 21 LLMs on \benchmark{}, revealing that most models experience significant performance drops as context length increases. Our further analysis shows that even state-of-the-art LLMs still have significant room for improvement in providing robust reasoning across different tasks. We have open-sourced \benchmark{} to support the comprehensive evaluation of LLMs' long-context reasoning capabilities.
}

\date{\today}
\correspondence{Zhan Ling at \email{z6ling@ucsd.edu}, Jiecao Chen at \email{jiecao.chen@bytedance.com}}
\checkdata[Dataset]{\url{https://huggingface.co/datasets/lz1bytedance/LongReason}}

\begin{document}

\maketitle


\section{Introduction}


\begin{table*}[t!]
\centering
\setlength{\tabcolsep}{0.6pt}
\resizebox{\textwidth}{!}{
\begin{tabular}{@{}l|c|ccccc@{}}
\toprule
\bf{Benchmark} &  
\begin{tabular}[c]{@{}c@{}}\hspace{0.5em} \bf{Avg} \hspace{0.5em} \\ \bf{Len} \end{tabular} & 
\begin{tabular}[c]{@{}c@{}}\hspace{0.5em} \bf{Light} \hspace{0.5em} \\ \bf{Human Effort} \end{tabular} & 
\begin{tabular}[c]{@{}c@{}}\hspace{0.5em} \bf{Realistic} \hspace{0.5em} \\ \bf{Tasks} \end{tabular}& 
\begin{tabular}[c]{@{}c@{}}\hspace{0.5em} \bf{Broad} \hspace{0.5em} \\  \bf{Tasks}  \end{tabular} & 
\begin{tabular}[c]{@{}c@{}}\hspace{0.5em} \bf{Controllable} \hspace{0.5em} \\ \bf{Context} \end{tabular}  \\
\midrule
{\bf{ZeroSCROLLS}}~\cite{zeroscrolls} & $\sim$10K & \xmark & \cmark & \cmark & \xmark \\
{\bf{L-Eval}}~\cite{leval} &  $\sim$8K & \xmark & \cmark & \cmark & \xmark  \\
{\bf{BAMBOO}}~\cite{dong2023bamboo} & $\sim$16K & \xmark & \cmark & \cmark & \xmark \\
{\bf{LongBench}}~\cite{bai2023longbench} & $\sim$8K &  \xmark & \cmark & \cmark & \xmark \\
{\bf{LooGLE}}~\cite{li2023loogle}  & $\sim$20K &  \xmark & \cmark & \cmark & \xmark \\
\midrule
{\bf{InfiniteBench}}~\cite{zhang2024infty}  & $\sim$200K & \xmark & \cmark & \cmark & \xmark \\
{\bf{Loong}}~\cite{wang2024leave} & $\sim$250K & \xmark & \cmark & \cmark & \xmark \\
\midrule
{\bf{Needle-in-a-haystack}}~\cite{needleinhaystack} & any & \cmark & \xmark & \xmark & \cmark \\
{\bf{RULER}}~\cite{hsieh2024ruler}  & any & \cmark & \xmark & \cmark & \cmark \\
\midrule
\bf{\benchmark{}} (Ours) & any & \cmark & \cmark & \cmark & \cmark \\
\bottomrule
\end{tabular}}
\caption{Comparison of \benchmark{} with other long-context benchmarks. \benchmark{} offers controllable context lengths and incorporating diverse and realistic tasks without the need for human annotation on long text. }
\label{tab:related_benchmarks}
\end{table*}

In recent years, large language models (LLMs)~\cite{openai2023gpt4, gemini, claude3, llama3-1, mistral, qwen2.5} have demonstrated remarkable advances in diverse natural
language processing tasks. The ability to comprehend and reason over long inputs is essential for downstream applications, including multi-turn conversations~\cite{tan2024peer}, document understanding~\cite{masry2024longfin} retrieval-augmented generation~\cite{yao2022react, xu2023retrieval}, and language agents~\cite{zhao2024longagent, zhang2024chain}. Meanwhile, extensive efforts in deep learning system~\cite{dao2022flashattention, dao2023flashattention, chen2023longlora, ratner2022parallel} research have been devoted to optimizing computational overhead to support increasing numbers of input tokens, which has led to growing attention on long-context LLMs. Now, both proprietary and open-source LLMs can support up to millions of input tokens~\cite{gemini, mistralnemo, glm4}.

However, despite the rapid development of long-context language models, benchmarks have lagged behind. One of the key challenges is dataset construction, as long-context question-answering data is relatively scarce on the internet. To address this, prevalent long-context benchmarks have utilized synthetic tasks like passkey retrieval~\cite{mohtashami2023landmark}, needle-in-a-haystack~(NIAH)~\cite{needleinhaystack, zhao2024longagent}, and variable tracking~\cite{hsieh2024ruler} to evaluate long-context LLMs. However, these tasks are often unrealistic and involve reasoning processes that differ significantly from those in real-world applications. Alternatively, some research efforts have involved human annotation of realistic questions and gold answers over one or multiple long documents~\cite{bamboo, wang2024leave, li2023loogle}. However, creating realistic long-context tasks from extensive texts is both challenging and time-consuming, even for human experts~\cite{wang2024leave}. This limitation restricts the expansion of datasets to accommodate arbitrary context lengths and the ability to support controllable context.
As shown in Table~\ref{tab:related_benchmarks}, existing benchmarks either rely on a limited number of synthetic tasks, demand significant human effort to read long contexts, or lacking controllable contexts and support for arbitrary context lengths. Furthermore, existing datasets~\cite{bamboo, wang2024leave, li2023loogle} often utilize documents from specific domains, such as financial reports or legal cases, as input, which can inherently limit the diversity of task categories. Consequently, they tend to focus on a narrow set of tasks, such as comparison or classification, rather than evaluating more complex and challenging tasks that require chain-of-thought reasoning. 

To address these challenges, we introduce a new long-context reasoning benchmark, \benchmark{}, featuring diverse and realistic reasoning tasks to assess the long-context reasoning abilities of LLMs. To create the dataset efficiently and effectively, we first had human annotators collect short reasoning questions from the internet, cleaning them to avoid data contamination and forming the seed dataset. This seed dataset contains reasoning questions with diverse patterns from three major task categories: reading comprehension, logical inference, and mathematical word problems. We chose to use multiple-choice problems for easy evaluation, avoiding the use of LLMs or inaccurate metrics like Rouge score and F1 to assess the correctness of reasoning. Then, we utilize an automatic pipeline that synthesizes multi-hop long-context reasoning questions from the collected short-context problems. To ensure quality, we leverage LLMs to automatically verify the generated questions, ensuring they retain the same logic as their shorter counterparts. Ultimately, we retain 794 questions that pass these checks. For each question, we can generate long-context versions of arbitrary lengths; however, since most existing models support contexts up to 128K tokens, we focus our evaluation within this limit. This synthetic pipeline supports converting one short reasoning question into different lengths, enabling fine-grained assessment of LLMs across various context lengths and reasoning tasks.

\begin{figure*}[t!]
    \centering
    \includegraphics[width=\textwidth]{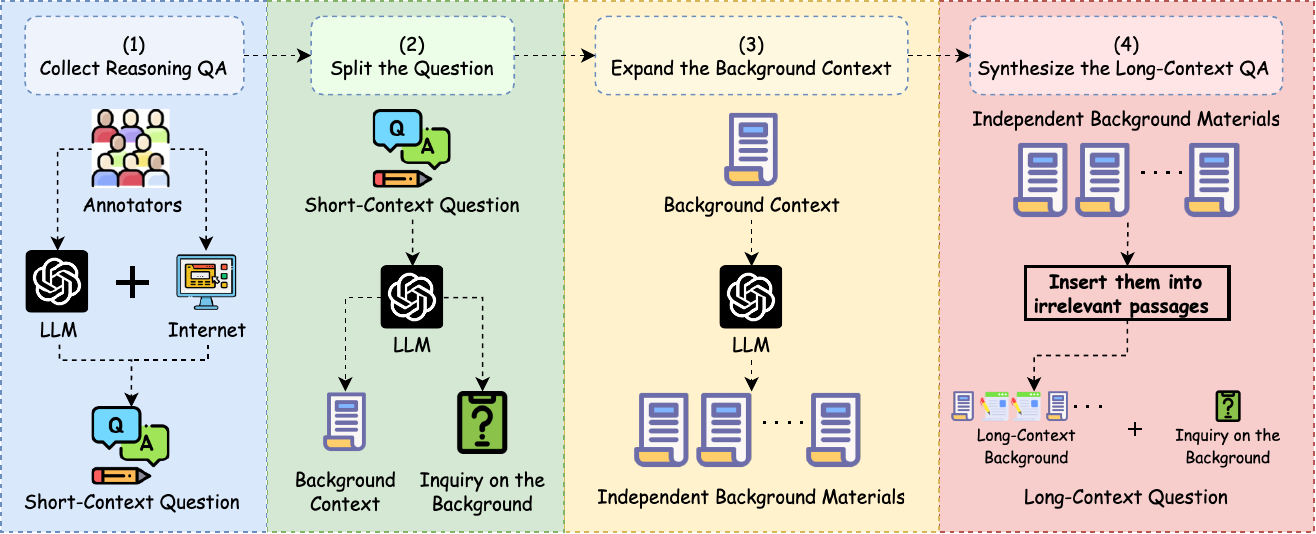}
    \caption{Overview of our pipeline for constructing \benchmark{}. Givem a short reasoning question $Q_{\text{short}}$, the pipeline first separates it into a background context $C_{\text{short}}$ and a final question $I$. Next, multiple paragraphs are synthesized from the background context $C_{\text{short}}$. These synthesized paragraphs are then embedded within irrelevant passages to create a long-context background. Finally, the constructed context is combined with the final question to generate the long-context reasoning question $Q_{\text{long}}$.}
    \label{fig:benchmark_pipeline}
\end{figure*}

To assess the current progress in the long-context reasoning abilities of existing LLMs, we evaluated 21 models of varying scales and architectures, sourced from both open-source and closed-source communities. While most of these models achieve near-saturated performance on previous synthetic tasks such as NIAH, nearly all exhibit significant performance degradation on \benchmark{} as the context length increases. Further analysis reveals that even state-of-the-art LLMs show varying degrees of performance decline across different task categories, underscoring the importance of evaluating diverse reasoning tasks to fully understand the long-context reasoning capabilities of LLMs.

\noindent Our key contributions are summarized as follows:
\begin{itemize}
    \item We present \benchmark{}, a new synthetic long-context reasoning benchmark that encompasses a diverse range of task categories and supports controllable context lengths.
    \item We propose an innovative synthesis algorithm that generates long-context reasoning questions from existing short questions, reducing the need for labor-intensive human annotation for long-context data.
    \item We perform an extensive analysis of current LLMs, benchmarking their performance in long-context reasoning and offering valuable insights to enhance long-context reasoning capabilities.
\end{itemize}

\section{Related Work}

\textbf{Long-Context Large Language Models}
Recent advancements in deep learning system have significantly propelled the development of long-context large language models~(LLMs). One of the key challenges in scaling these models is the quadratic time and space complexity inherent in computing self-attention over long sequences. To mitigate this computational burden, efficient self-attention algorithms~\cite{dao2022flashattention, dao2023flashattention, liu2023ring} have been introduced, reducing memory overhead, and novel training methods~\cite{li2021sequence, chen2023longlora} facilitate the training of these long-context models. As Rotary Position Embedding (RoPE)~\cite{su2024roformer} is widely used for positional encoding in many open-source models~\cite{llama3-1, qwen2.5, mistrallarge2}, recent research~\cite{pi, xiong2023effective, peng2023yarn, liu20242, ding2024longrope, pose} has focused on adapting RoPE from pre-trained short-context models to effectively handle longer sequences. Moreover, new architectures~\cite{mamba, rwkv, bulatov2022recurrent, rmt, botev2024recurrentgemma} have been developed to efficiently process long-context inputs. Consequently, state-of-the-art language models~\cite{openai2024gpt4o, openai2024gpt4omini, reid2024gemini, anthropic2024claud35sonnet, llama3-1, mistrallarge2, qwen2.5, glm4} now support context windows ranging from 128K to millions of tokens, enabling the exploration of reasoning abilities over extensive contexts with LLMs.

\begin{figure*}[t!]
    \centering    \includegraphics[width=\textwidth]{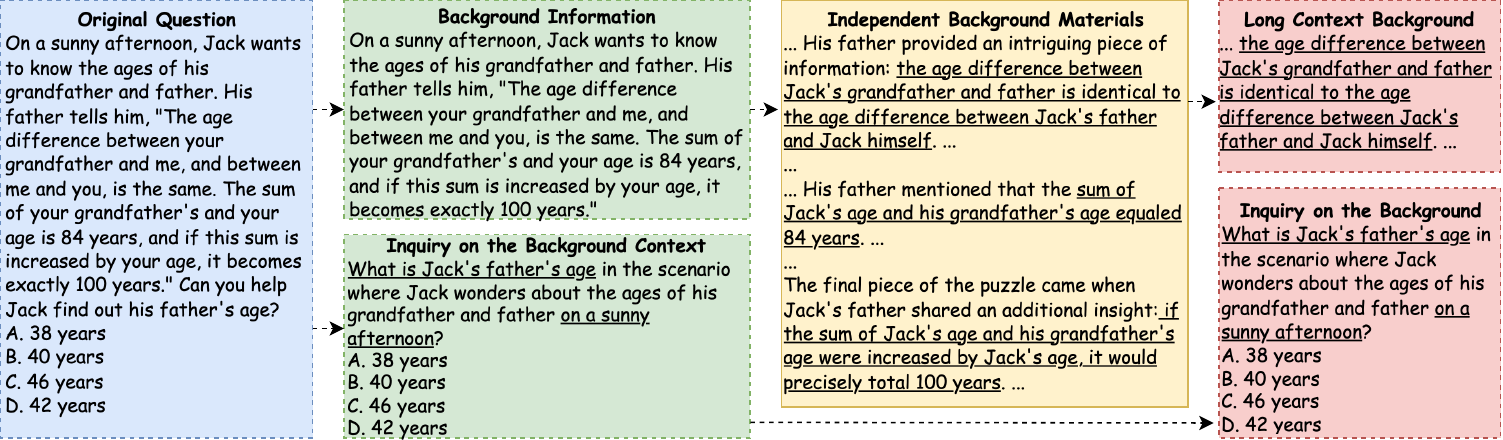}
    \caption{An illustrative example in \benchmark{}. The original question is first decomposed into a separate background passage and an inquiry based on it. The inquiry includes keywords such as “Jack’s father’s age” and a time reference like “on a sunny afternoon” from the background passage, ensuring a clear connection to the passage. Subsequently, the background passage is expanded into multiple independent materials while preserving these key keywords. Finally, these independent materials are combined with some unrelated passages to create the final long-context reasoning question.}
    \label{fig:benchmark_example}
\end{figure*}

\textbf{Long-Context Benchmarks}
As the context window of current LLMs expands rapidly, numerous benchmarks have been proposed to evaluate their capabilities. In early benchmarks such as ZeroSCROLLS~\cite{zeroscrolls}, L-Eval~\cite{leval}, BAMBOO~\cite{bamboo}, LongBench~\cite{bai2023longbench}, and LooGLE~\cite{li2023loogle}, the average input length remains under 25K tokens, which is far shorter than the context window size supported by existing LLMs. Recently, some research has begun to explore using synthetic datasets, which can support controllable context lengths, to evaluate the long-context abilities of LLMs. Needle-in-a-Haystack and its variants~\cite{needleinhaystack, zhao2024longagent} primarily evaluate retrieval abilities by inserting relevant information into extensive irrelevant corpora and testing the LLMs’ capacity to extract it. Additionally, RULER~\cite{hsieh2024ruler} constructs synthetic tasks based on code-like flexible configurations to assess LLM performance over long contexts. While synthetic tasks can support the evaluation of arbitrarily long contexts, they are limited in scope, focusing on a narrow set of tasks and failing to comprehensively evaluate the reasoning abilities of LLMs in realistic scenarios. Other benchmarks like InfiniteBench~\cite{zhang2024infty} and Loong~\cite{wang2024leave} use human annotations to create questions from given long texts, which contain more diverse tasks but are both time-consuming and costly. Our proposed benchmark, \benchmark{}, focuses on evaluating the long-context reasoning abilities of LLMs, which are created automatically from short reasoning questions without heavy human effort in reading the long context. We conduct a detailed comparison with existing benchmarks in Table~\ref{tab:related_benchmarks}.

\section{Our Benchmark: \benchmark{}}
\label{sec:benchmark}
In this section, we provide a detailed overview of \benchmark{}, our synthetic long-context reasoning benchmark. This includes the problem formulation, the dataset construction process, and an analysis of the statistics of \benchmark{}.

\subsection{Long-context Reasoning Question Construction via Context Expansion}
\label{sec:construction}

\textbf{Problem Formulation}
The primary goal of \benchmark{} is to assess the long-context reasoning abilities of LLMs. To achieve this, we first define the reasoning task as follows: Given a reasoning question $Q$, LLMs to need reason over $Q$ to produce a reasoning chain $S$ that leads to the final answer $A$. In this work, the focus is on scenarios where the question $Q$ can be divided into a background context $C$ and a final inquiry $I$ based on that context, denoted as $Q=(C, I)$. In \benchmark{}, the context $C$ can be long, comprising multiple paragraphs from diverse sources, while only a small subset of the information in the context $C$ is directly relevant to answering $I$. To simplify evaluation, \benchmark{} employs close-ended multiple-choice questions for $I$. The dataset construction begins with a set of questions $\textbf{Q}_{\text{short}}$, consisting of questions $Q_{\text{short}}$ with relatively short question statements. For each $Q_{\text{short}}$, our proposed context expansion pipeline utilizes LLMs to automatically generate a long-context version of the question, $Q_{\text{long}} = (C_{\text{long}}, I)$. The detailed construction pipeline is illustrated in Figure~\ref{fig:benchmark_pipeline}.

\noindent \textbf{Short-Context Reasoning Question Collection} 
We begin by asking human annotators to create a dataset  $\textbf{Q}_{\text{short}}$, comprising short questions $Q_{\text{short}}$ across various domains and diverse task categories. Annotators collect example questions from the internet and utilize an LLM to refine these questions, ensuring they are free from data contamination. To ensure that each short question require reasoning, we prompt an LLM to evaluate the number of reasoning steps in its corresponding ground-truth reasoning chain, denoted as $\bar{S}$. We include only those questions that require at least two reasoning steps to arrive at the final answer in \benchmark{}, thereby filtering out straightforward common-sense problems that lack significant reasoning depth.

\noindent \textbf{Automatic Short-Context Reasoning Question Decomposition with LLMs}
For each short reasoning question $Q_{\text{short}}$, we prompt an LLM to decompose the question into a background context $C_{\text{short}}$ and an inquiry $I$. This decomposition needs to ensure that the final inquiry $I$ is clearly linked to the background context $C_{\text{short}}$ , enabling the LLM to relate them and answer the inquiry based on the context. To have the better performance, we prompt the LLM to perform the decomposition in a chain-of-thought manner. Specifically, the LLM first extracts key elements such as keywords, time, main characters, and event names from the original short question and incorporates them into both the background context $C_{\text{short}}$ and the final inquiry $I$ during the decomposition process. To ensure the quality of the decomposition, we introduce a self-verification stage after generating the decomposed question. We ask the LLM to verify whether the decomposed question, $Q_{\text{decomposed}}=(C_{\text{short}}, I)$, retains the same meaning as the original question $Q_{\text{short}}$. For each question, we use a sampling temperature of 0.7 and generate up to 5 decompositions with the LLM. We retain only the decomposition that successfully passes the self-verification process conducted by the LLM. In our experiments, we found that over 99.34\% of questions could be successfully decomposed within 5 samples, demonstrating the effectiveness of our question decomposition pipeline.

\noindent \textbf{Automatic Background Context Decomposition with LLMs} To evaluate the ability to aggregate key information and reason across different positions within a long context, we further decompose the background context $C_{\text{short}}$ in the question $Q_{\text{decomposed}}$ into multiple information pieces. Specifically, we use an LLM to first analyze all key information points within $C_{\text{short}}$ and then, for each information point, generate an independent and complete passage $C'$. These generated passages retain certain keywords similar to those used during the question decomposition stage, ensuring that all passages are closely related to the final inquiry $I$. This process results in $C_{\text{expanded}} = (C'_1, C'_2, \cdots)$ , where the passages are coherent and can be correctly associated with the final inquiry $I$. To ensure the quality of the expanded context, we introduce a self-verification stage. After generating the expanded question $Q_{\text{expanded}} = (C_{\text{expanded}}, I)$, we prompt the LLM to verify whether $Q_{\text{expanded}}$ retains the same meaning as the original question $Q_{\text{short}}$. For the background in the each question, we use a sampling temperature of 0.7 and generate up to 5 decompositions with the LLM. Only the decompositions that successfully pass the self-verification process are retained. In our experiments, we observed that over 94.67\% of the background contexts were successfully decomposed within 5 samples.

\noindent \textbf{Automatic Background Decomposition with LLMs} To evaluate the ability to aggregate key information and reason across different positions within a long context, we further decompose the background context $C_{\text{short}}$ in the question $Q_{\text{decomposed}}$ into multiple information pieces. Specifically, we use an LLM to first analyze all key information points within $C_{\text{short}}$ and then, for each information point, generate an independent and complete passage $C'$. These generated passages retain certain keywords similar to those used during the question decomposition stage, ensuring that all passages are closely related to the final inquiry $I$. This process results in $\bar{C}_{\text{expanded}} = (C_e^1, C_e^2, \cdots)$ , where the passages are coherent and can be correctly associated with the final inquiry $I$. To ensure the quality of the expanded context, we introduce a self-verification stage. After generating the expanded question $Q_{\text{expanded}} = (\bar{C}_{\text{expanded}}, I)$, we prompt the LLM to verify whether $Q_{\text{expanded}}$ retains the same meaning as the original question $Q_{\text{short}}$. For the background context in the each question, we use a sampling temperature of 0.7 and generate up to 5 decompositions with the LLM. Only the decompositions that successfully pass the self-verification process are retained. In our experiments, we observed that over 94.67\% of the background contexts were successfully decomposed within 5 samples.

\begin{figure}[t]
    \centering
    \includegraphics[width=0.55\linewidth]{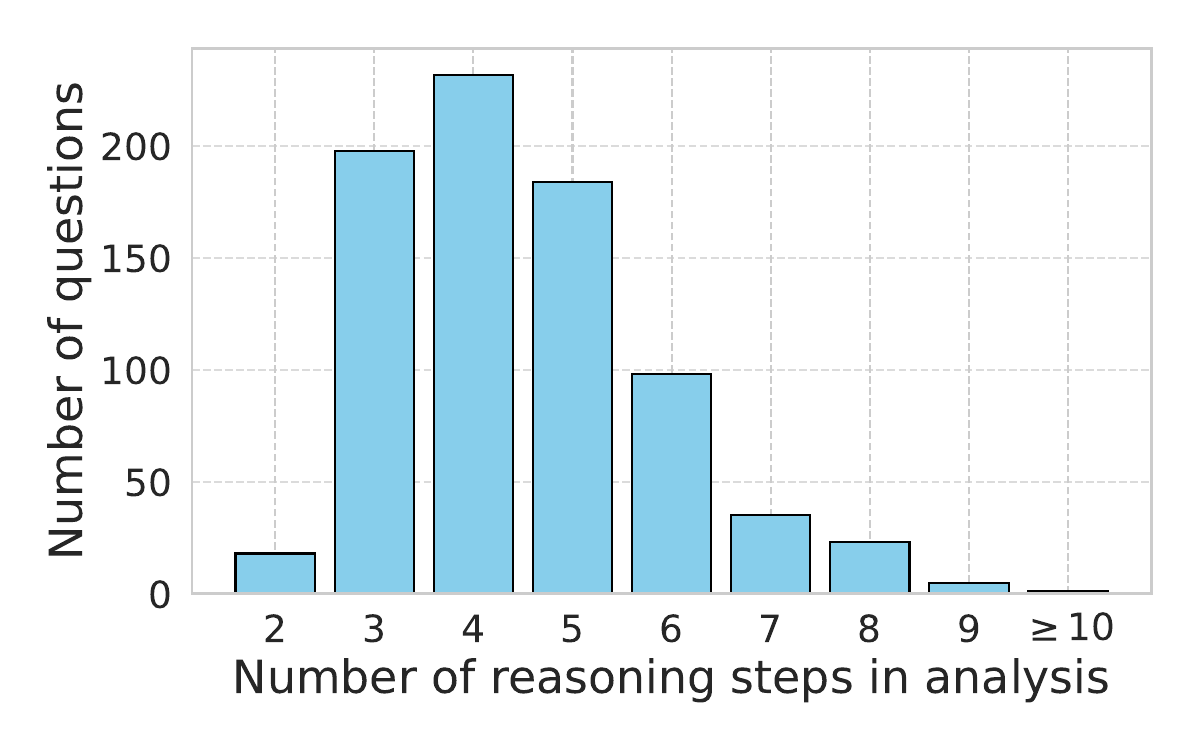}
    \caption{The number of reasoning steps in the ground-truth analysis for questions in \benchmark{}.}
    \label{fig:step_distribution}
\end{figure}

\begin{table*}[t]
\centering
\resizebox{\textwidth}{!}{
\begin{tabular}{@{}l|c|cc|cccccc}
\toprule
 \bf Models & \bf Length & \bf Q-O & \bf Q-E & \bf 8K & \bf 16K & \bf 32K & \bf 64K & \bf 128K & \bf{Avg.}  \\
\midrule
Random & - & 25.21  & 25.21 & 25.21 & 25.21 & 25.21 & 25.21 & 25.21 & 25.21  \\
\shline
 \multicolumn{10}{c}{\textit{closed-source models}}  \\
 Gemini-1.5 Pro & -   & \textbf{90.42} & 84.11 & \textbf{77.81} & \textbf{79.70} & \textbf{77.81} & \textbf{78.94} & \textbf{78.81} & \textbf{78.56} \\
 Gemini-1.5 Flash & -   & 90.16 & 80.20 & 75.91 & 76.29 & 75.79 & 75.66 & 76.92 & 75.91 \\
  GPT-4o     & -   & \textbf{90.42} & 85.62 & 77.30 & 76.80 & 74.91 & 74.02 & 73.39 & 75.76 \\
 GPT-4o mini & -   & 79.95 & 74.40 & 68.73 & 66.83 & 65.45 & 62.67 & 61.66 & 65.92 \\
  Claude-3.5 Sonnet   & -   & 84.36 & 78.18 & 73.01 & 70.11 & 68.47 & 68.22 & 65.95 & 69.95 \\
 Claude-3.5 Haiku    & -   & 77.05 & 71.75 & 64.44 & 64.44 & 63.93 & 60.03 & 59.90 & 63.21 \\
 \shline
 \multicolumn{10}{c}{\textit{open-source models}}  \\
 Llama-3.1-70B      & 128K  & 80.83 & 74.27 & 68.22 & 66.46 & 61.16 & 63.30 & 48.30 & 64.78 \\
 Llama-3.1-8B       & 128K  & 58.13 & 57.12 & 53.47 & 51.20 & 51.45 & 49.94 & 46.53 & 51.52 \\
 Mistral Large 2 & 128K  & 83.73 & 81.97 & 72.89 & 70.11 & 64.69 & 52.46 & 0.00~\footnotemark[0]  & 65.04 \\
Mixtral 8x22B & 64K & 64.69 & 63.30 & 50.95 & 52.21 & 49.31 & 48.68 & -    & 50.29 \\
 Mistral Nemo  & 1M & 56.12 & 52.96 & 50.57 & 43.00 & 42.37 & 38.21 & 29.51 & 43.54 \\
 Mistral Small & 32K & 50.32 & 64.94 & 56.75 & 50.32 & 37.70 & -    & -    & 48.26 \\
 Mistral-7B    & 32K & 41.61 & 40.86 & 44.77 & 43.25 & 42.75 & -    & -    & 43.59 \\
 Qwen2.5-72B        & 128K  & \textbf{89.16} & \textbf{85.75} & 76.67 & \textbf{77.43} & \textbf{74.27} & \textbf{74.53} & \textbf{69.48} & \textbf{75.72} \\
 Qwen2.5-32B        & 128K  & 84.24 & 81.59 & \textbf{78.44} & 74.91 & 72.76 & 71.75 & 67.34 & 74.46 \\
 Qwen2.5-14B        & 128K  & 84.11 & 76.17 & 71.88 & 70.87 & 68.10 & 66.20 & 62.30 & 69.26 \\
 Qwen2.5-7B         & 128K  & 76.42 & 73.01 & 66.33 & 62.42 & 62.17 & 58.76 & 54.22 & 62.42 \\
 Qwen2.5-3B         & 32K & 61.29 & 58.76 & 49.81 & 49.56 & 45.65 & -    & -    & 48.34 \\
  Phi-3.5-MoE        & 128K  & 65.32 & 66.46 & 48.42 & 56.37 & 53.72 & 48.80 & 49.56 & 51.83 \\
 Phi-3.5-mini       & 128K  & 55.99 & 60.53 & 50.69 & 48.80 & 49.81 & 45.40 & 24.97 & 48.68 \\
 glm-4-9b     & 128K  & 59.90 & 63.43 & 48.68 & 46.15 & 44.14 & 38.97 & 39.60 & 44.48 \\
\bottomrule
\end{tabular}
}
\caption{Performance~(\%) of selected LLMs on \benchmark{}. All the scores are computed by averaging the accuracy across 794 questions in \benchmark{}. Q-O represents the performance of the original short question $Q_{\text{short}}$, and Q-E denotes the performance of the expanded question $Q_{\text{expanded}}$ mention in Section~\ref{sec:construction}. For long-context questions, the final inquiry is placed after the background context, positioning it at the end of the context. The average score~(Avg.) represents the mean performance across context lengths spanning from 8K to 128K.}
\label{tab:main_result}
\end{table*}


\noindent \textbf{Long-Context Reasoning Question Construction Through Context Expansion}
Finally, we construct the long-context version of each question by embedding each passage $C_e^i$ from the expanded context $\bar{C}{\text{expanded}}$ at random positions within a set of irrelevant passages $\bar{C}{\text{irrelevant}}$, forming the final long-context reasoning questions. To create $\bar{C}{\text{irrelevant}}$ , we first collect passages from the Pile~\cite{gao2020pile} and use an LLM to rewrite each passage to minimize stylistic differences between the synthesized background passages and the irrelevant passages. These rewritten passages are then compiled to form the set of irrelevant passages $\bar{C}_{\text{irrelevant}}$. In \benchmark{}, GPT-4 is used for all data synthesis and self-verification. For each question, we evaluate multiple versions of the synthesized question for comparison, including the original question $Q_{\text{short}}$, the expanded version  $Q_{\text{expanded}}$, and long-context versions with context lengths ranging from $8K$ to $128K$. Furthermore, similar to NIAH~\cite{needleinhaystack}, our pipeline is capable of generating reasoning questions with even longer contexts by incorporating additional irrelevant information.

\subsection{The Statistics of \benchmark{}}
\footnotetext[0]{Mistral Large 2 generates an empty response when the question context length reaches 128K.}


\benchmark{} comprises 794 multiple-choice reasoning questions encompassing diverse reasoning patterns across three task categories: 280 reading comprehension questions, 347 logical inference questions, and 167 mathematical word problems. We only keep the questions that require at least 2 reasoning steps, the reasoning steps of the questions range from 2 to more than 10 reasoning steps. The average reasoning steps of the questions is 4.47. More detailed statistics of the number of the reasoning steps are shown in Figure~\ref{fig:step_distribution}.

\section{Exerperiments \& Results}
\label{sec:exp}

We conduct a comprehensive set of experiments to evaluate a broad set of LLMs using \benchmark{}. In this section, we present the experimental setup, main results, and additional analysis.

\subsection{Experimental setup}
\noindent\textbf{Models \& Inference Setup} We select a set of representative LLMs that support long context windows, including 6 closed-source models from 3 model families~(GPT, Gemini and Claude) and 15 open-source models spanning a wide range of model sizes (3B to 123B) and claimed context lengths (8K to 2M). Detailed information about these models can be found in Appendix~\ref{app:model_info}. For open-source models, we utilize vLLM~\cite{vllm}, which enables efficient KV cache memory management during inference time. All inferences are performed using \textit{bfloat16} precision on 8 NVIDIA A100 GPUs with greedy decoding~(temperature=0).

\noindent\textbf{Evaluation setup} We evaluate all models on \benchmark{}, which comprises 794 questions, each featuring multiple variations, including the original version, expanded versions, and long-context versions with context lengths of 8K, 16K, 32K, 64K, and 128K. Each input is constructed using a predefined zero-shot chain-of-thought template that combines the background context, followed by the corresponding final inquiry. Detailed information about the prompt template is provided in Appendix~\ref{app:prompt}. To assess the reasoning performance of the LLMs, we extract the predicted choice by identifying the first character  following the phrase “the answer is” and compare it to the ground-truth option for accuracy.

\subsection{Main Results}
\begin{figure}[t]
  \centering
  \includegraphics[width=0.5\linewidth]{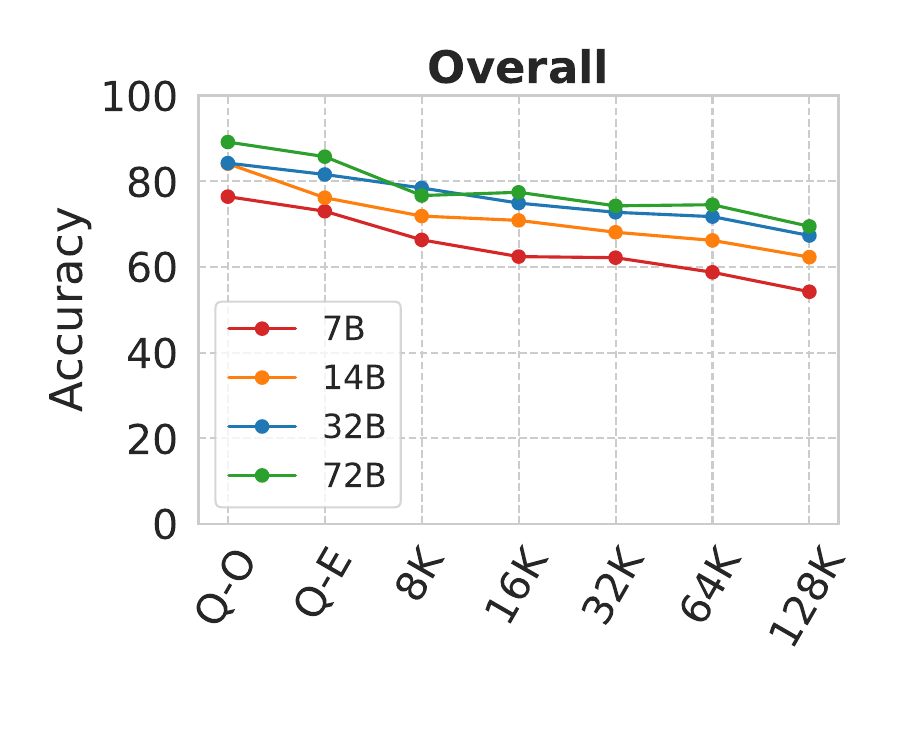}
  \vspace{-3em}
  \caption{Performance of the Qwen2.5 series on \benchmark{}, with model sizes ranging from 7B to 72B.}
  \label{fig:compare_model_size}
\end{figure}

The results of 21 LLMs are presented in Table~\ref{tab:main_result}. From the table, we first observe a significant performance drop across nearly all models when evaluated on $Q_{\text{expanded}}$ compared to $Q_{\text{short}} $. To ensure this decline is not caused by the quality of the synthetic questions, we manually examine 20 failure cases from Gemini-1.5 Pro, where correct answers on $Q_{\text{short}}$ turn incorrect on $Q_{\text{expanded}}$. Only 3 cases involve ambiguity or errors introduced by context expansion. Similarly, when comparing $Q_{\text{expanded}}$ to $Q_{8K}$ , a large performance drop persists. Among 20 failure cases from Gemini-1.5 Pro where correct answers on $Q_{8K}$ turn incorrect on $Q_{\text{expanded}}$, only 2 cases are affected by added irrelevant information. For long-context reasoning performance, Gemini-1.5 Pro outperforms all other closed-source models, exhibiting negligible performance drop when extending the context length from 8K to 128K. In contrast, the long-context reasoning capabilities of open-source LLMs lag behind those of the most advanced closed-source models in \benchmark{}. For example, the best-performing open-source model, Qwen2.5-72B, experiences a significant performance drop~(5.05\%) when the input context length increases from 64K to 128K. Furthermore, a comparison of Qwen2.5 models of different sizes, shown in Figure~\ref{fig:compare_model_size}, reveals that performance declines at a similar rate across all model sizes as context length increases. Smaller models perform worse overall, primarily due to their weaker reasoning abilities, even in shorter-context scenarios.

\subsection{Further Analysis}
We conduct further analysis on \benchmark{} to provide a deeper understanding of the long-context reasoning performance of existing LLMs.

\noindent\textbf{Does the position of the final inquiry influence model performance?} As shown in Table \ref{tab:ablation_question_position}, the performance of state-of-the-art language models is highly sensitive to the position of the final inquiry. Although Gemini-1.5 Pro demonstrates excellent long-context reasoning performance when the final inquiry is placed after the background context, it still struggles when the inquiry is positioned at the beginning of the input, before the background context. Meanwhile, GPT-4o demonstrates similar performance in both cases, particularly when the context length is short. However, as the input length increases, GPT-4o’s performance declines significantly for questions with the final inquiry is placed before the background context.
\begin{table*}[t]
\centering
\resizebox{\textwidth}{!}{
\begin{tabular}{l|cc|cc|cc|cc|cc}
\toprule
 \multirow{2}{*}{\bf{Model}} & \multicolumn{2}{c}{\textbf{8K}} & \multicolumn{2}{c}{\textbf{16K}} &  \multicolumn{2}{c}{\textbf{32K}} &  \multicolumn{2}{c}{\textbf{64K}} & \multicolumn{2}{c}{\textbf{128K}} \\
 \cmidrule{2-11}
 & I-L & I-F & I-L & I-F & I-L & I-F & I-L & I-F & I-L & I-F\\
\shline
 GPT-4o         & 77.30     & \textbf{75.41}      & 76.80      & \textbf{72.89}       & 74.91      & \textbf{69.10 }      & 74.02      & 65.32       & 73.39       & 65.95        \\
 Gemini-1.5 Pro & \textbf{77.81}     & 68.22      & \textbf{79.70}      & 70.37       & \textbf{77.81}      & 68.60       & \textbf{78.94}      & \textbf{66.96}       & \textbf{78.81}       & \textbf{66.71}        \\
 Claude-3.5 Sonnet   & 73.01     & 68.60      & 70.11      & 67.84       & 68.47      & 66.46       & 68.22      & 64.69       & 65.95       & 66.20        \\
\bottomrule   
\end{tabular}
}
\caption{Ablation study on the position of the final inquiry for selected models evaluated at context lengths ranging from 8K to 128K. I-L represents questions where the final inquiry is placed after the background context, while I-F represents questions where the inquiry is placed before the background context.}
\label{tab:ablation_question_position}
\end{table*}

\begin{figure*}[h!]
    \centering
    \includegraphics[width=\textwidth, keepaspectratio]{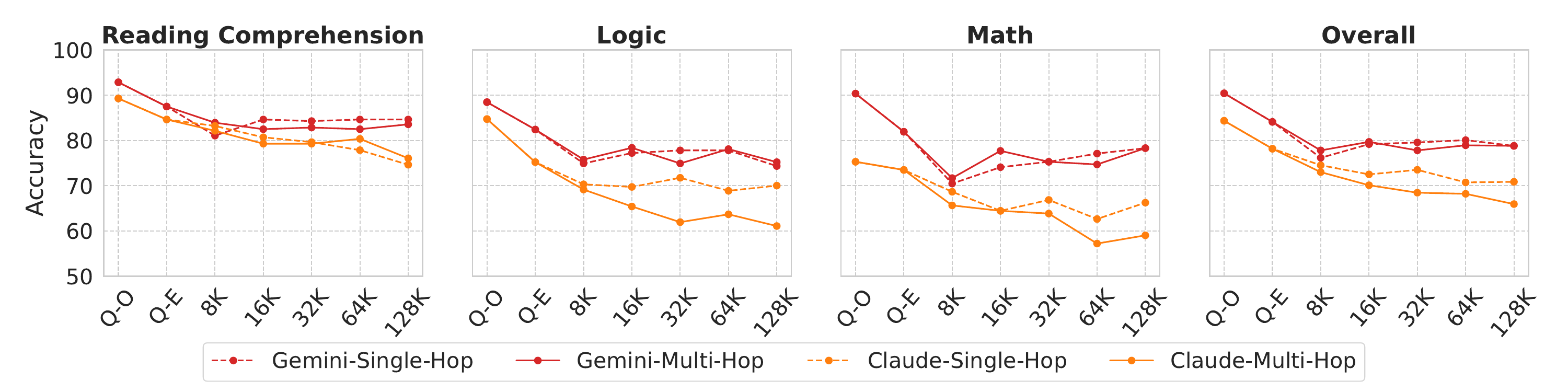}
    \caption{Comparison of the long-context reasoning performance between Gemini-1.5 Pro and Claude 3.5-Sonnet across different task categories. In the figure, the dotted line represents the single-hop version of the synthesized questions, where all clues are placed together in the context. The solid line represents the multi-hop version, which is the standard format used in \benchmark{}, where clues are distributed separately throughout the context.}
    \label{fig:compare_gemini_claude}
\end{figure*}

\noindent\textbf{Do LLMs have similar long-context reasoning performance over different tasks and clue placement in \benchmark{}?}
As shown in Figure~\ref{fig:compare_gemini_claude}, both Gemini-1.5 Pro and Claude 3.5 demonstrate strong long-context reasoning performance on reading comprehension problems. However, for logic and math problems, Claude 3.5 significantly underperforms compared to Gemini. Additionally, we observe that for these problem types, Claude 3.5 shows much lower performance when the clues are distributed separately throughout the context, compared to when the clues are grouped together.


\begin{figure}[h!]
    \centering
    \includegraphics[width=0.5\linewidth, keepaspectratio]{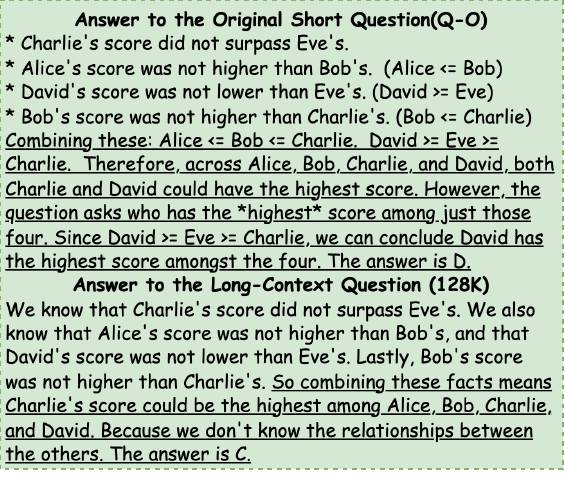}
    \caption{An example illustrating how Gemini-1.5 Pro provides incorrect reasoning for a long-context question but correct reasoning for the original short question. The key difference in reasoning is underlined in the figure.}
    \label{fig:failure_case_1}
\end{figure}

\noindent\textbf{Error Cases Analysis}
We analyze 20 randomly sampled error cases from Gemini-1.5 Pro on questions with a 128K context. Among these, we find only 3 instances where the errors are caused by missing critical information in the background context during reasoning, while the remaining cases were attributed to reasoning errors. A detailed example is provided in Figure~\ref{fig:failure_case_1}.

\section{Conclusion and Limitations}
In this work, we introduce \benchmark{}, a synthetic reasoning benchmark designed to evaluate the long-context reasoning capabilities of large language models (LLMs). Using \benchmark{}, we evaluate the long-context reasoning performance of 21 LLMs across context sizes ranging from 8K to 128K. Our experiments and analyses reveal that existing LLMs still have significant room for improvement in delivering robust long-context reasoning. Additionally, several limitations of \benchmark{} remain, as discussed below.

\textbf{Lack of evaluation for complex reasoning} Current \benchmark{} primarily focuses on evaluating reasoning questions that require only a few reasoning steps. However, this is insufficient to fully understand the performance of LLMs when dealing with challenging problems that demand many reasoning steps over a long context.

\textbf{Lack of evaluation for tasks requiring full context} Similar to most existing work, \benchmark{} focuses on tasks that do not require understanding the entire contexts for finishing the tasks. All the questions in \benchmark{} are derived from short reasoning problems that can be solved by examining only a small portion of the context. 




\bibliographystyle{plainnat}
\bibliography{main}

\clearpage

\beginappendix

\section{Model Information}
\label{app:model_info}
We select in total 21 large language models~(LLMs) for evaluation and analysis. We only include the aligned models including 6 clouse-source models like GPT-4o, Gemini-1.5, and Claude-3.5 and also 17 open-source models with dense and MoE architectures like Llama and Mixtral using LongReason.

\begin{table*}[h]
    \centering
    \small
    \resizebox{\linewidth}{!}{
    \begin{tabular}{@{}lcccl@{}}
    \toprule
    Model & Aligned & Size & Context Length & Huggingface~\citep{huggingface} / API \\
    \midrule
    GPT-4o~\citep{openai2024gpt4o} & \ding{51} & - & 128K & \texttt{gpt-4o-2024-08-06} \\
    GPT-4o-mini~\citep{openai2024gpt4omini} & \ding{51} & - & 128K & \texttt{gpt-4o-mini-2024-07-18} \\
    Gemini-1.5-Pro~\citep{gemini} & \ding{51} & - & 2M & \texttt{gemini-1.5-pro-002} \\
    Gemini-1.5-Flash~\citep{gemini} & \ding{51} & - & 2M & \texttt{gemini-1.5-flash-002} \\
    Claude-3.5-Sonnet\citep{anthropic2024claud35sonnet} & \ding{51} & - & 200K & \texttt{claude-3-5-sonnet-20240620} \\
    Claude-3.5-Haiku\citep{anthropic2024claud35sonnet} & \ding{51} & - & 200K & \texttt{claude-3-5-haiku-20241022} \\
    \midrule
    Mistral-Large2~\citep{mistrallarge2} & \ding{51} & 123B & 128K & mistralai/Mistral-Large-Instruct-2407 \\
    Mixtral-8$\times$22B~\citep{jiang2024mixtral} & \ding{51} & 39B/8$\times$22B & 64K & mistralai/Mixtral-8x22B-Instruct-v0.1 \\
    Mistral-Small~\citep{mistraltec} & \ding{51} & 22B & 32K & mistralai/Mistral-Small-Instruct-2409 \\
    Mistral-Nemo~\citep{mistralnemo} & \ding{51} & 12B & 1M & mistralai/Mistral-Nemo-Instruct-2407 \\
    Mistral-7B~\citep{mistral} & \ding{51} & 7B & 32K & mistralai/Mistral-7B-Instruct-v0.3 \\

    Llama3.1~\citep{llama3-1} & \ding{51} & 70B & 128K & meta-llama/Meta-Llama-3.1-70B-Instruct \\
    Llama3.1~\citep{llama3-1} & \ding{51} & 8B & 128K & meta-llama/Meta-Llama-3.1-8B-Instruct \\

    Qwen2.5~\citep{qwen2.5} & \ding{51} & 72B & 128K & Qwen/Qwen2.5-72B-Instruct \\
    Qwen2.5~\citep{qwen2.5} & \ding{51} & 32B & 128K & Qwen/Qwen2.5-32B-Instruct \\
    Qwen2.5~\citep{qwen2.5} & \ding{51} & 14B & 128K & Qwen/Qwen2.5-14B-Instruct \\
    Qwen2.5~\citep{qwen2.5} & \ding{51} & 7B & 128K & Qwen/Qwen2.5-7B-Instruct \\
    Qwen2.5~\citep{qwen2.5} & \ding{51} & 3B & 32K & Qwen/Qwen2.5-3B-Instruct \\

    GLM4-9B~\citep{glm4} & \ding{51} & 9B & 128K & THUDM/glm-4-9b-chat \\


    Phi3.5-MoE~\citep{phi3} & \ding{51} & 6.6B/16$\times$3.8B & 128K & microsoft/Phi-3.5-MoE-instruct \\
    Phi3.5-mini~\citep{phi3} & \ding{51} & 14B & 128K & microsoft/Phi-3.5-mini-instruct \\
    
    \bottomrule
    \end{tabular}
}
    \caption{Information of evaluated and analyzed models in~\benchmark.}
    \label{app:models}
\end{table*}

\section{Hyperparameters for \benchmark{} Construction}
In \benchmark{}, we utilize gpt-4o-2024-08-06 to synthesize our dataset, with the total cost of creating the datasets being under \$200.

\section{Human Annotator}
To create the short reasoning questions with human annotations, we trained five researchers from our research group following the requirements outlined in the paper.

\clearpage
\section{Prompts}
\label{app:prompt}

\begin{table*}[h]
\centering
\scriptsize
\begin{tabular}{l} 
\toprule
\textbf{\#\# Question}\\
\textit{\{question\}}\\
\\
\textbf{\#\# Instruction}\\
Please break down the question below into a background passage and a question related to this background passage \\ according to the following requirements:\\
1. The background passage should contain a complete context and retain as much of the original wording as possible, \\ but do not add any extra facts, as this may affect the correct answer to the final question. Do not include options in the \\ background passage!\\
2. Since the background passage will be mixed in with a large amount of other materials, the question about the \\ background passage needs to have a clear signal pointing to this specific background passage. This signal is preferably \\ the name of a character, location, or event from the background passage, but it should not include the core content of \\ the story and should contain as little information from the background passage as possible.\\
3. Do not use any pronouns in the question about the background passage.\\
4. If the original question contains options, the question about the background passage should also include options. \\ When writing options, list each option on a new line, using letters such as A, B, C, D, and E as labels. Do not include \\ any blank options from the original question.\\
5. Please write the background passage and the question about the background passage in English.\\
\\
\textbf{\#\# Answer Format}\\
\textbf{\#\#\# Analysis}\\
Please provide a brief analysis of how to perfectly break down this question into a background passage and a question \\ related to this passage, particularly focusing on what signal should be used in the question to refer to the background \\ passage.\\
\\
\textbf{\#\#\# Broken Down Question}\\
\textbf{\#\#\#\# Background Passage}\\
A background passage that meets my requirements. Please use English and do not include any additional information.\\
\textbf{\#\#\#\# Question About the Background Passage}\\
A question about the background passage that meets my requirements. Please use English and do not include any \\ additional information.\\
\\
\textbf{\#\#\# Analysis of the Question About the Background Passage}\\
Briefly analyze the question about the background passage and judge whether, if the background passage is mixed in with \\ a large amount of other materials, you would be able to find the related background passage after reading the "question \\ about the background passage."\\
\\
\textbf{\#\#\# Judge Whether the Question About the Background Passage Meets the Requirements}\\
Yes or No. Do not provide any additional information.\\
\\
\textbf{\#\#\# Analyze Whether the Broken Down Question Matches the Original Question in Meaning}\\
Briefly analyze whether the broken down question is consistent in meaning with the original question.\\
\\
\textbf{\#\#\# Judge Whether the Broken Down Question Matches the Original Question in Meaning}\\
Yes or No. Do not provide any additional information.\\
\\
\textbf{\#\# Respond to My Instructions According to the Above Format}\\
\bottomrule
\end{tabular}
\caption{Zero-shot prompt for separating a short reasoning quesiton into a background context and a final inquiry.} 
\label{tab:prompt_split_question}
\end{table*}

\begin{table*}
	\centering
	\scriptsize
	\begin{tabular}{l} 
		\toprule
\textbf{\#\# Material}\\
\textit{\{context\}}\\
\\
\textbf{\#\# Question about the Material}\\
\textit{\{final\_question\}}\\
\\
\textbf{\#\# Instructions}\\
Please expand the above material into multiple independent paragraphs with around 200 words in English, while meeting \\ the following requirements.
1. Ensure that every key piece of information from the material appears in one paragraph of \\ the expanded text. Try to place the key information in the middle of the expanded paragraphs.\\
2. The expanded material need to avoid introducing additional knowledge, reasoning, or any content that might influence \\ the answer to the "Question about the material."\\
3. The expanded text should clearly relate to the "Question about the material." Please include hints or references to the \\ question within each paragraph to maintain this connection. However, do not use words like "question" or "query" \\ explicitly in the expanded text.\\
4. Each paragraph should be a standalone piece of text, comprehensible without needing to refer to other paragraphs. \\ Minimize the use of pronouns, particularly those referring to other paragraphs.\\
5. Do not reference or imply any possible answer choices that might be part of the "Question about the material."\\
6. The style of the expanded text should match the specified target genre: \textit{\{target\_genre\}}.\\
\\
\textbf{\#\# Response Format}\\
\textbf{\#\#\# Analysis}\\
Please analyze how to appropriately add background information to expand the material into multiple  independent \\ paragraphs based on the given requirements. Additionally, assess whether the provided material can be easily divided into \\ multiple paragraphs. If not please only provide only one paragraph in the expanded material.\\
\textbf{\#\#\# Expanded material}\\
Present the expanded material as a series of independent paragraphs that meet the above requirements. Add the index \\ like "1." (do not use any format here) at the beginning of each paragraph (starting from 1), use English, and do not \\ include any extra information.\\
\\
\textbf{\#\# Respond to My Instructions According to the Above Format}\\
		\bottomrule
	\end{tabular}
	\caption{Zero-shot prompt for expanding the given short context into several independent passages.} 
\label{tab:prompt_split_context}
\end{table*}

\begin{table*}
	\centering
	\scriptsize
	\begin{tabular}{l} 
		\toprule
\textbf{\#\# Original material}\\
\textit{\{context\}}\\
\\
\textbf{\#\# Question about the material}\\
\textbf{\{final\_question\}}\\
\\
\textbf{\#\#\# Expanded material}\\
\textit{\{expanded\_context\}}\\
\\
\#\# Instructions\\
Please compare the expanded material with the original material and answer the following questions:\\
1. Does the expanded material contain all the key information from the original material?\\
2. Does the expanded material contain information that will affect the answer to the question?\\
3. Do you think all the paragraphs in the expanded material are related to the main topic/character of the question?\\
\\
\textbf{\#\# Response Format}\\
\textbf{\#\#\# Analysis}\\
Please analyze the expanded material and compare it with the original material. Then, combine the question and analyze the \\ three questions above.\\
\textbf{\#\#\# Does the expanded material contain all the key information from the original material based on} \\ \textbf{ the analysis?}\\
Yes or No. Do not provide any additional information.\\
\textbf{\#\#\# Does the expanded material contain information that will affect the answer to the question based} \\ \textbf{on the analysis?}\\
Yes or No. Do not provide any additional information.\\
\textbf{\#\#\# Are all the paragraphs in the expanded material are related to the main topic/character of the } \\ \textbf{question based on the analysis?}\\
Yes or No. Do not provide any additional information.\\
\\
\textbf{\#\# Respond to My Instructions According to the Above Format}\\
		\bottomrule
	\end{tabular}
	\caption{Zero-shot prompt for assessing the quality of the synthesized background context.} 
\label{tab:prompt_verify_splitted_background}
\end{table*}

\begin{table*}
	\centering
	\scriptsize
	\begin{tabular}{l} 
		\toprule
\textbf{\#\#\# Question}\\
\textit{\{question\}}\\
\\
Please answer the above question based on the background information!\\
\\
\textbf{\#\#\# Answer}\\
Please analyze step by step, and provide the final answer in the last line using "The answer is" + option (represented by ABCDE)!\\
		\bottomrule
	\end{tabular}
	\caption{Zero-shot chain-of-thought prompt for answering the given question.} 
\label{tab:prompt_cot_question}
\end{table*}

\begin{table*}
	\centering
	\scriptsize
	\begin{tabular}{l} 
		\toprule
\textbf{\#\#\# Background Information}\\
\textit{\{context\}}\\
\\
\textbf{\#\#\# Question} about the Background Information\\
\textit{\{final\_question\}}\\
\\
Please answer the above question based on the background information!\\
\\
\textbf{\#\#\# Answer}\\
Please analyze step by step, and provide the final answer in the last line using "The answer is" + option (represented by ABCDE)!\\
		\bottomrule
	\end{tabular}
	\caption{Zero-shot prompt chain-of-thought for answering the given question based on the background context.} 
\label{tab:prompt_cot_question_with_context}
\end{table*}

\end{document}